\documentclass{article}

\usepackage[english]{babel}
\usepackage[letterpaper,top=2cm,bottom=2cm,left=3cm,right=3cm,marginparwidth=1.75cm]{geometry}
\usepackage{amsmath}
\usepackage{amssymb,amsthm}
\usepackage{algorithm}
\usepackage{algpseudocode}
\usepackage{graphicx}
\usepackage{appendix}
\usepackage{authblk}

\usepackage{graphicx}
\newtheorem{theorem}{Theorem}

\title{Horizontal and Vertical Federated Causal Structure Learning via Higher-order Cumulants}

\author[1] {Wei Chen}
\author[1] {Wanyang Gu}
\author[1] {Linjun Peng} 
\author[1,2] {Ruichu Cai}
\author[1,3] {Zhifeng Hao}
\author[4,5] {Kun Zhang}

\affil[1]{School of Computer Science, Guangdong University of Technology, Guangzhou, China}
\affil[2]{Peng Cheng Laboratory, Shenzhen, China}
\affil[3]{College of Science, Shantou University, Shantou, China}
\affil[4]{Department of Philosophy, Carnegie Mellon University, Pittsburgh, PA, United States}
\affil[5]{Mohamed bin Zayed University of Artificial Intelligence, Abu Dhabi, United Arab Emirates}

\date{}
\begin{document}
\maketitle

\begin{abstract}

Federated causal discovery aims to uncover the causal relationships between entities while protecting data privacy, which has significant importance and numerous applications in real-world scenarios. Existing federated causal structure learning methods primarily focus on horizontal federated settings. However, in practical situations, different clients may not necessarily contain data on the same variables. In a single client, the incomplete set of variables can easily lead to spurious causal relationships, thereby affecting the information transmitted to other clients. To address this issue, we comprehensively consider causal structure learning methods under both horizontal and vertical federated settings. We provide the identification theories and methods for learning causal structure in the horizontal and vertical federal setting via higher-order cumulants. Specifically, we first aggregate higher-order cumulant information from all participating clients to construct global cumulant estimates. These global estimates are then used for recursive source identification, ultimately yielding a global causal strength matrix. Our approach not only enables the reconstruction of causal graphs but also facilitates the estimation of causal strength coefficients. Our algorithm demonstrates superior performance in experiments conducted on both synthetic data and real-world data.

\end{abstract}

\section{Introduction}

The purpose of causal discovery is to reveal the causal relationships between variables or events from observational data, also known as causal structure learning. Current machine learning techniques, particularly deep learning models, exhibit limited interpretability. However, causal models have excellent interpretability. Developing effective causal discovery methods to enhance interpretability plays a crucial role, and it has been widely applied in various fields including social science \cite{1}, artificial intelligence \cite{2}, systems biology \cite{3} and medicine \cite{4}, to enable inference and analysis of events. 

With the increasing frequency of data leakage incidents and heightened public awareness regarding data security, a growing number of individuals and organizations are hesitant to share their private datasets. This trend presents significant challenges to traditional causal discovery methods, which rely on centralized data environments \cite{5}. As a result, federated causal discovery aimed at uncovering causal relationships from decentralized data sources has attracted considerable attention and has been extensively studied by researchers \cite{6}.

Most research in federated causal discovery (FCD) focuses on the horizontal federated learning. These methods operate solely at the sample level and account for client data heterogeneity. However, in vertically partitioned or mixed federated causal discovery scenarios where clients possess partially overlapping feature sets and no single client holds the complete variable set existing constraint-based and score-based FCD methods face significant limitations. Locally, clients may be unable to identify complete separating sets, leading to inaccurate conditional independence (CI) tests. This introduces spurious edges in local causal graphs. Subsequent global aggregation on the server then propagates these errors, resulting in numerous redundant edges in the global causal structure.

Similarly, constraint-based methods to FCD (e.g., FedC$^2$SL \cite{7} and FedCDH \cite{8}), which implement federated conditional independence tests (FCIT), also typically rely on finding valid separating sets. Consequently, they cannot be directly applied to vertically federated settings. Furthermore, due to the fundamental definition of the cumulative quantity and its inherent sample decomposition characteristics, this method can also be effectively applied in horizontal federation scenarios. In practice, the scenario of a hybrid federation appears in many fields, such as collaborative medical diagnosis \cite{28}, recommendation systems \cite{29}, and graph learning \cite{30}, where the data collected by different clients have different and possibly overlapping features and sample IDs \cite{31}. Specific examples include multiple hospitals, each equipped with different diagnostic equipment. As a result, the collected features will be different. Moreover, these hospitals may be unable to diagnose the same patient. To tackle the above issues, we aim to provide a method for federated causal structure learning in both horizontal and vertical settings. Interestingly, we find that, higher-order cumulants rely solely on the joint distribution of the relevant variables and are not influenced by the absence of variables \cite{27}. In the linear non-Gaussian case, higher-order cumulants can capture the information of the absence of variables from existing variables within each client \cite{cai2023causal,chen2024identification}. These motivate us to propose identifiability theories for horizontal and vertical federated causal structure learning via higher-order cumulants. Based on these, we propose an algorithm for federated identification of sources based on high-order cumulants. Our main contributions can be outlined as follows:
\begin{itemize}
    \item The proposed FedISHC method can identify the entire directed acyclic causal structure from horizontal and vertical mixed federal scenarios.
    \item The FedISHC method we proposed ingeniously transforms this into eliminating the influence of the discovered source variables at the cumulant level, thereby finding the global causal order.
    \item In the existing FCIT, the incomplete set of variables on the client side leads to inaccurate FCIT results. In contrast, high-order cumulants only depend on the joint distribution of the involved variables, and their calculation is not affected by the lack of data from other variables.
\end{itemize}

\section{Related Work}
In recent years, many FCD methods have been proposed , which can be broadly categorized into three classes: continuous optimization-based methods, score-based methods, and constraint-based methods. 

Continuous optimization-based methods, such as NOTEARS-ADMM \cite{9} applies the NOTEARS algorithm to a distributed setting by using the distributed ADMM optimization method. Additionally, the authors proposed two approaches: NOTEARS-ADMM for the discovery of linear causal relationships, and NOTEARS-MLP-ADMM for the discovery of nonlinear causal relationships. Building on this foundation, FedDAG \cite{5} is used to discover directed acyclic graph structures from distributed data under the assumption of additive noise model. It separates causal structure learning and causal mechanism approximation through a two-level structure in local models. FedCausal \cite{10} proposed a global optimization formula to aggregate the local causal graph parameters from client data, while also restricting the acyclicity of the global graph. Enhanced the ability of the FCD method to handle more complex data.

Score-based methods, DARLS \cite{11} using distributed annealing on regularized likelihood scores to learn a unified causal graph from decentralized data with privacy protection and theoretical guarantees of converging to an oracle solution. PERI \cite{12} using distributed min-max regret optimization with theoretical guarantees of consistency and $\epsilon$-differential privacy, demonstrating superior performance in discovering causal networks from decentralized data.

Constraint-based methods, most of the FCD methods are discussed in this section, FedPC \cite{13} adapts the PC algorithm to the federated learning setting through layer-wise skeleton aggregation and consistent separation set identification strategies, but it is only applicable to homogeneous data. To address the issue of FedPC's performance degradation when local samples are limited, FedECD \cite{14} employs Bootstrap technology to resample local data and generate multiple local datasets. This process enhances the robustness of causal skeleton learning through a two-layer aggregation strategy. FedCSL \cite{15} addresses the scalability and accuracy limitations of existing federated causal structure learning algorithms like FedPC by employing a federated local-to-global learning strategy and a novel weighted aggregation strategy, which overcomes the scalability and accuracy issues encountered by existing federated CSL algorithms. FedACD \cite{16} addresses the sample quality heterogeneity issue in federated causal discovery by adaptively selecting and sending only the optimal causal relationships learned under high-quality variable spaces from each client to the server, while masking those learned under low-quality spaces, thereby achieving accurate FCD. FedC$^2$SL \cite{7} is the first work to apply the federated conditional independent test protocol to federated causal structure learning, and it is tolerant of client heterogeneity. To further adapt to the heterogeneity of client data, FedCDH \cite{8} employs a surrogate variable to address data heterogeneity across clients, enabling the effective handling of both heterogeneous data and arbitrary causal models. Furthermore, it introduces a federated conditional independence test (FCIT) for skeleton discovery and a federated independent change principle (FICP) for determining causal directions.

Causal Discovery from Multiple Data Sets. ION \cite{17} integrates locally learned causal structures from multiple datasets with overlapping variables to output globally consistent causal models. COmbINE \cite{18} infers causal structure from multiple interventions on overlapping variables via constraint-based SAT encoding. CD-MiNi \cite{19} stacks data via overcomplete ICA, enforces shared causal structure across datasets, and resolves indeterminacies using non-Gaussian noise ratios at overlapping variables. CDUIOV \cite{20} resolves entangled inconsistencies from unknown interventions and overlapping variables to learn causal structures across domains.

\section{Notations and Assumptions}

\subsection{Assumption}
\newtheorem{assumption}{\bf Assumption}
\begin{assumption}\label{asp:1}
Let X be the union of variables from all K datasets, i.e., $X=\bigcup_{k=1}^{K}X^k=(x_1,x_2,...,x_d)$, with a total of $d$ variables. For any $x_i$ and $x_j$,  there must exist at least one client among them.
\end{assumption}

\subsection{Notations}

Assume that we have $K$ clients, each with a local dataset. Let $X^k~(k=\{1,…,K\})$ be the set of observed variables in the dataset of the $k$-th client. Each dataset has $|X^k|=d^k$ variables. The datasets may contain different sets of variables. Let $X$ be the union of variables from all $K$ datasets, i.e., $X=\bigcup_{k=1}^{K}X^k$, with a total of $d$ variables. Each client has $n_k$ samples. For any given variable $x_i$, it is not guaranteed that all clients possess this variable. Consequently, a direct aggregation of the sample sizes across all clients is not feasible.

A causal structure over $X$ is often represented using a causal directed acyclic graph (DAG) \cite{21}.
In a causal DAG, if there is a direct edge $x_i \rightarrow x_j~(i, j\in\{1, 2,\dots,d\})$,~$x_i$ is a direct cause (parent) of $x_j$, and $x_j$ is a direct effect (child) of $x_i$ \cite{22}.

Definition 1 (Cumulant \cite{23}). Let $X = (x_1,x_2,...,x_d)$ be a
random vector of length d. The $k$-th order cumulant tensor of
$X$ is defined as a $d×· · ·×d$ ($k$ times) table, $C^{(k)}$, whose entry at position $(i_1, · · · , i_k)$ is
$$C^{(k)}_{i_1,···,i_k} = Cum(x_{i_1},\dots,x_{i_k})=\sum_{(D_1,\dots,D_h)}(-1)^{h-1}(h-1)! \mathbb{E}\left[\prod_{j \in D_i} x_j\right]\ldots\mathbb{E}\left[\prod_{j \in D_i} x_j\right]$$
where the sum is taken over all partitions $(D_1,\dots,D_h)$ of the set ${i_1,\dots,i_k}$.

For convenience, we use $C_{m}(x_i)$ to denote the cumulant $Cum(\underbrace{x_i,\dots}_{m~\textrm{times}})$, and use $C_{m,n}(x_i, x_j)$ to denote the joint cumulant $Cum(\underbrace{x_i,\dots}_{m~\textrm{times}},\underbrace{x_j,\dots}_{n~\textrm{times}})$. For example, $C_3(x_i)$ represents $Cum(x_i, x_i, x_i)$, $C_{1,2}(x_i, x_j)$ represents $Cum(x_i, x_j, x_j)$.

Note that the first-order cumulant of a random variable $x_i$ corresponds to its mean, while the second-order cumulant is equivalent to the variance. These two statistics form the core characteristics of the Gaussian distribution. However, when dealing with non-Gaussian distributions, relying solely on the mean and variance is insufficient to fully characterize the data distribution. Here, high-order cumulants play a crucial role: the third-order cumulant quantifies the skewness of $x_i$'s distribution, reflecting the degree of deviation from symmetry, while the fourth-order cumulant measures the kurtosis of $x_i$, indicating the tendency of data to cluster around the center or exhibit extreme values.The third-order cumulant (skewness) and fourth-order cumulant (kurtosis) are the important and widely used
measures of non-Gaussianity \cite{24}. In essence, the probability density function of a Gaussian distribution can be completely described by first and second-order statistics alone. For complex non-Gaussian distributions, however, high-order cumulants serve as key tools to accurately capture the high-order characteristics of the data and reveal the essence of the distribution.

Definition 2 (Linear Non-Gaussian Acyclic Model \cite{25}). Suppose the causal graph over X is a directed acyclic graph (DAG). Let $X=(x_1,x_2,...,x_d)$ denote a $d$-dimensional vector of observed variables. Then, $x$ follows the data generation process:
\begin{equation}
X=BX+E\label{1}
\end{equation}
where $E=(e_1,e_2,...,e_d)^\mathrm{T}$ is the vector of independant Non-Gaussian noise terms, and B is the causal strength matrix describing the influences of the variables $X$ on each other. $b_{ij}$ represents the element in the $i$-th row and $j$-th column of matrix B, indicating the causal influence of variable $x_j$ on variable $x_i$. the causal strength $b_{ij}$, for any $i$ and $j$, is fixed in all local datasets. otherwise, B is a strictly lower triangular matrix if one knew a causal ordering $S$ of the variables. 

\begin{figure}[ht]
    \centering
    \includegraphics[width=1\textwidth]{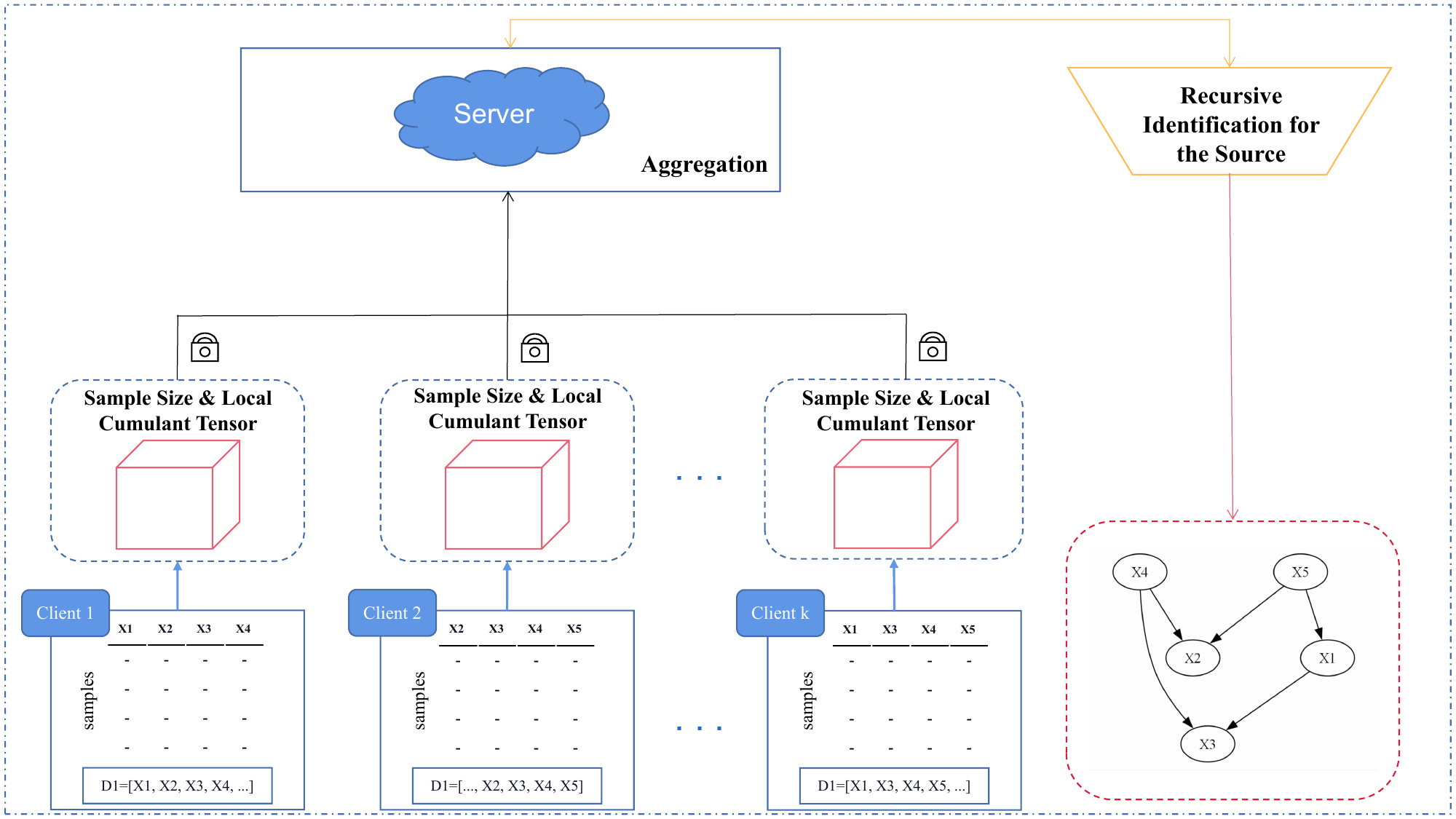}     
    \caption{Overall framework of FedISHC. First, the client sends its sample size and local cumulant tensor to the server in order to construct the global cumulant tensor. The global cumulant tensor will be aggregated on the server to obtain the result. Based on this global cumulant tensor, the server uses recursive identification to trace the algorithm and ultimately obtains the global causal graph.}
    \label{fig:FedISHC}
\end{figure}

\section{Method}
As shown in Fig. \ref{fig:FedISHC}, the proposed FedISHC consists of five steps: calculate the local cumulant matrix of the clients, aggregate all local cumulant matrices for the service, obtain the current source variable using the global cumulant information, eliminate the influence of the source variable and achieve recursive identification of the source variable and obtain the global causal adjacency matrix B. 

Step 1: Each client calculates the local cumulant matrix from its own local data set and sends it to the server.

Cumulants, which are sample-wise decomposable and independent of the variable set. Thus, the overall cumulant reasonably equals the sum of individual local cumulants, making it suitable for federated learning. Consequently, we first compute the local cumulant on each client. For client $k$, it computes all third-order cumulants between any variable and the others, constructs a local cumulant matrix, and sends it to the server. The cumulant matrix is in the form of:
$$    
Cum^{k}=
\left(                
  \begin{array} {ccccccc}
     C_3(x_1) & C_{1,2}(x_1,x_2) & ...& C_{1,2}(x_1,x_{d_k}) & C_{2,1}(x_1,x_2) & ... & C_{2,1}(x_1,x_{d_k})\\
    C_{1,2}(x_2,x_1) & C_3(x_2)  & ... & C_{1,2}(x_2,x_{d_k}) & C_{2,1}(x_2,x_1) & ... & C_{2,1}(x_2,x_{d_k})\\
    ... & ... & ... & ... & ... & ... & ...\\
    C_{1,2}(x_{d_k},x_1) & C_{1,2}(x_{d_k},x_2) & ... & C_3(x_{d_k}) & C_{2,1}(x_{d_k},x_1) & ... & C_{2,1}(x_{d_k},x_{d_{k-1}})\\
  \end{array}
\right)                 
$$
The dimension of the cumulant matrix is $d^k \times (2d^k-1)$, For example, the cumulant matrix form of client $1$ with $3$ variables is: $$Cum^{1}=
\left(                
  \begin{array}{ccccc}
    C_3(x_1) & C_{1,2}(x_1,x_2) & C_{1,2}(x_1,x_3) & C_{2,1}(x_1,x_2) & C_{2,1}(x_1,x_3)\\
    C_{1,2}(x_2,x_1) & C_3(x_2) & C_{1,2}(x_2,x_3) & C_{2,1}(x_2,x_1) & C_{2,1}(x_2,x_3)\\
    C_{1,2}(x_3,x_1) & C_{1,2}(x_3,x_2) & C_3(x_3) & C_{2,1}(x_3,x_1) & C_{2,1}(x_3,x_2)\\
  \end{array}
\right)     $$
Step 2: The server aggregates all the local cumulant matrices of the clients to form a global cumulant matrix.

First, the global variable set $X$ is obtained by taking the union of all client variable sets (i.e., $X=\bigcup_{k=1}^{K}X^k$). Then, establish the mapping between the local variable set and the global variable set. Aggregate according to the corresponding positions of the mapping vector. For instance, the value at position $i,j$ in the global cumulant matrix is calculated using equation \eqref{2}:

\begin{equation}
Cum_{ij}=\frac{n_1}{n_1+...+n_k}Cum_{ij}^{1}+...+\frac{n_k}{n_1+...+n_k}Cum_{ij}^{k}(\{1,...,k\} \in I_{ij}~(I_{ij}=\{ k |  i\in O^k \cap j\in O^k \}))\label{2}
\end{equation}
where $O^k$ is the variable index set of the $X^k$ observation variable set, $I_{ij}$ is the set of all client indices that simultaneously contain variable indices $i$ and $j$. According to the sample-wise decomposable property of cumulants, the equation \eqref{2} holds.

Step 3: Using the global cumulant matrix information, calculate equation \eqref{3} for each variable with other variables to find the initial source variables.

\begin{equation}
\tau_{ij}=|C_3(x_i)C_{1,2}(x_i,x_j)-C_{2,1}(x_i,x_j)C_{1,2}(x_j,x_i)|\label{3}
\end{equation}

As higher-order cumulants can extract essential information from data, and the joint cumulants of variables indicate linear relationship properties, we present the following intuition. Suppose $x_i$ is a direct causal of $x_j$. Some (joint) cumulants of $x_i$ and $x_j$ can be expressed as follows:
\begin{equation} \label{4}
\begin{aligned}
&C_3(x_i) = C_3(e_i) \\
&C_3(x_j) = a_{ji}^3C_3(e_i)+C_3(e_j) \\
&C_{1,2}(x_i,x_j)=C_{2,1}(x_j,x_i)=a_{ji}^2C_3(e_i) \\
&C_{2,1}(x_i,x_j)=C_{1,2}(x_j,x_i)=a_{ji}C_3(e_i) \\
\end{aligned}
\end{equation}
Then we can observe an interesting conclusion that $\tau_{ij}=C_3(x_i)C_{1,2}(x_i,x_j)-C_{2,1}(x_i,x_j)C_{1,2}(x_j,x_i)=0$ and $\tau_{ji}=C_3(x_j)C_{1,2}(x_j,x_i)-C_{2,1}(x_j,x_i)C_{1,2}(x_i,x_j)=a_{ji}C_3(e_i)C_3(e_j)\neq 0$, this is because there is an extra term $e_j$ in $x_j$ that cannot be eliminated, which is induced by the result variable. Then we extended this conclusion to the case involving multiple variables, and proposed Theorem \ref{thm:1}.

Only the source variable satisfies Theorem \ref{thm:1}. At this moment, our most intuitive thought is to use it as a standard for identifying source variables within the framework of the DirectLiNGAM algorithm \cite{32}, but this is not feasible. Figure \ref{fig:FedISHC} shows that we currently have three clients. The variable set of client $1$ is $(x_1, x_2, x_3,x_4)$, the variable set of client $2$ is $(x_2, x_3, x_4,x_5)$, and the variable set of client $3$ is $(x_1, x_3, x_4,x_5)$. At this point, we assume that the current source variable being identified is $x_1$. We then broadcast this information to the clients. The clients use residual regression to eliminate the influence of the source variable. But client $2$ is unable to perform this operation, which will have an impact on the subsequent updates. Therefore, We ingeniously transformed it into a method that eliminates the influence of the discovered source variables at the level of cumulant. 

\begin{theorem} \label{thm:1}
Assume that the variables $X$ are generated by LiNGAM. Then $x_s \in X$ is the source variable, if and only if
$$x_s=\sum_{s\in O,~j\in O\textbackslash \{s\}} \tau_{sj}=0,$$where $O=\bigcup_{k=1}^K O^k$. 
\end{theorem}

The proof is provided in Appendix A1. Theorem \ref{thm:1} guarantees
that if for $x_s$, $\sum_{s\in \mathrm{U},~j\in \mathrm{O}\textbackslash \{s\}} \tau_{sj}=0$ for other variables, then $x_s$ is the source variable. When there are multiple source variables, they are called sibling source variables. The relationship between variables of the same level must be independent. Therefore, we use equation \eqref{3} to obtain the current source variable.

Step 4: Update the global cumulant matrix using equations \eqref{5},\eqref{6} and \eqref{7}, and recursively search for the current source variable.

First, after identifying the current source variable through Theorem 1, we need to eliminate the influence of the source variable on its descendant nodes. We use equations \eqref{5}, \eqref{6} and \eqref{7} to update the global cumulant matrix in order to eliminate the influence of the discovered source variables at the cumulant level. Therefore, we propose Theorem \ref{thm:2} to ensure the correctness of our approach of eliminating the influence of the discovered source variables at the cumulant level.

\begin{theorem} \label{thm:2}
For any three variables $x_i$, $x_j$, and $x_k$, if $x_i$ is the source variable, regardless of the structure of $x_j$ and $x_k$, the equations \eqref{5}, \eqref{6} and \eqref{7} will hold true, thereby eliminating the influence of $x_i$.
\end{theorem}
The proof is provided in Appendix A2. Theorem 2 ensures that we can eliminate the influence of the source variable on its descendant variables at the cumulant level.

\begin{equation}
C^{'}_3(x_j)=C_3(x_j)-\alpha^{3}C_3(x_i) \label{5}
\end{equation}
\begin{equation}
C^{'}_{1, 2}(x_j, x_k) = C_{1, 2}(x_j, x_k)-\alpha \lambda^2 C_3(x_i)\label{6}
\end{equation}
\begin{equation}
C_{2, 1}^{'}(x_j, x_k)= C_{2, 1}(x_j, x_k)-\alpha^2 \lambda C_3(x_i)\label{7}
\end{equation}
where $\alpha =C_{2, 1}(x_i, x_j)/C_3(x_i),\lambda = C_{2, 1}(x_i, x_k)/C_3(x_i)$ are respectively the Mixed influence coefficients of $x_i$ on $x_j$ and $x_k$.

\begin{figure}[ht]
    \centering
    \includegraphics[width=0.6\textwidth]{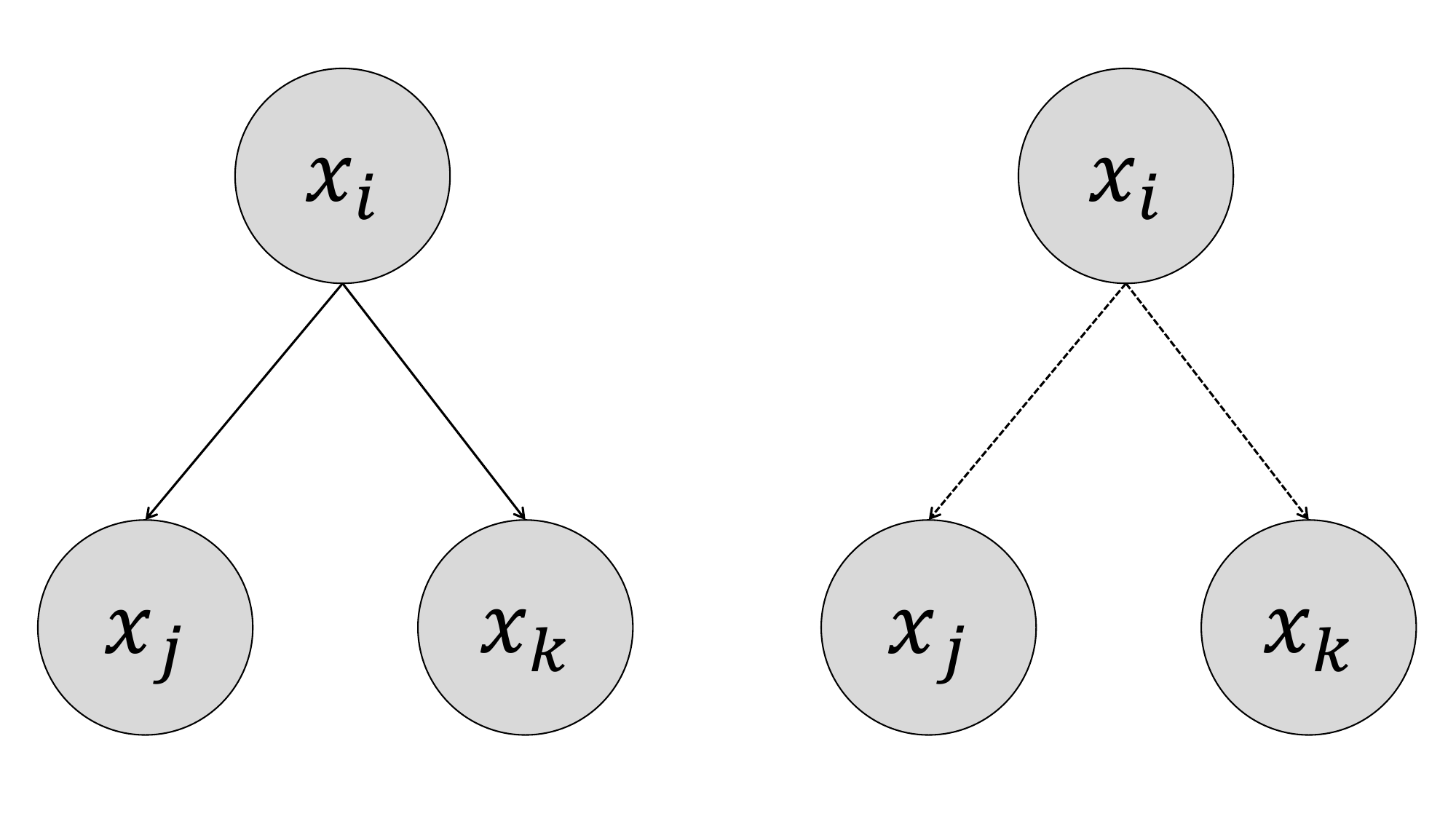}     
    \caption{The left figure shows that $x_i$, $x_j$, and $x_k$ form a common structure, while the right figure indicates that $x_i$, $x_j$, and $x_k$ are independent of each other.}
    \label{fig:LatentConfounder}
\end{figure}

Step 5: The algorithm outputs the global causal order, the updated global cumulant matrix, and the causal strength matrix $B$.

Throughout the algorithm's execution, we compute the influence coefficient $\alpha = C_{2,1}(x_s, x_j)/ C_{3}(x_s)$ for the source variable $x_s$ and each other variable $x_j$. This value of $\alpha$ is recorded in the $(s, j)$ position of the causal strength matrix $B$. Consequently, our algorithm simultaneously estimates the causal strength matrix $B$ while determining the causal order.







In practical, the value of our parameter $\tau$ is typically not exactly zero, and the underlying distribution of $\tau$ remains unknown. non-parametric methods such as the sign test or permutation test can be employed to assess whether $\tau$ can be considered as zero. Consequently, instead of transmitting the local cumulant matrix, the client sends the local cumulant tensor to the server. This is the reason why the client in Figure \ref{fig:FedISHC} actually sends a cumulant tensor.

\section{Privacy and costs analysis}
\subsection{PRIVACY ANALYSIS}
Initially, local cumulant matrix $\{Cum^k\}$ are learned in the respective FL clients, which require the raw data of each client, and then are shared to the server for obtaining the global cumulant matrix $Cum$. Next, The server utilizes the global cumulant matrix as the input for Algorithm 1, which is then executed recursively. Upon completion, Algorithm 1 produces the global causal order, the updated global cumulant matrix, and the causal strength matrix $B$. As a result, FedISHC exchanges higher-order statistics information rather than raw data, protecting data privacy to a certain extent. To further avoid data privacy leakage, secure multi-party computation can be implemented using Shamir's secret sharing scheme \cite{26}. In this approach, each client securely distributes secret shares of its local cumulant matrix to other participants. The clients then collaboratively reconstruct the aggregated matrix through distributed computation. Alternatively, homomorphic encryption techniques may be employed \cite{27}. Each client encrypts its local cumulant matrix using homomorphic encryption and transmits the encrypted data to the server. The server then performs aggregation operations directly on the homomorphically encrypted matrices. Subsequent decryption yields the aggregated result in plaintext.

\subsection{COMMUNICATION COSTS}
Assume the number of observed variables in client $k$ is $d^k$ and the number of integrated variables in the server is $d$. For each client $k$, Just send a $d^k \times (2d^k-1)$ matrix, and a value $n_k$, which requires $O(d^k \times (2d^k-1)+1)$. Moreover, in practice, an test needs to be conducted, suppose the number of tests is $N$, For each client $k$, Just send a $d^k \times (2d^k-1) \times N \times 1$ tensor, and a value $n_k$, which requires $O(d^k \times (2d^k-1) \times N \times 1+1)$.

\section{Extensions to More General Cases}
In this section, we demonstrate that the situation can be easily extended to the Linear Gaussian Model. Suppose there are three variables $x$, $y$, and $z$:

\begin{equation} \label{8}
\rho_{xy|z}=\frac{\rho_{xy}-\rho_{xz}\rho_{zy}}{\sqrt{(1-\rho_{xz}^2)(1-\rho_{zy}^2)}}
\end{equation}

The detailed proof process of Equation \eqref{8} is presented in Appendix A3. The partial correlation coefficient can be calculated using equation \eqref{8}, thereby extending our application scope to the case of linear Gaussian. However, in the case of linear Gaussian, the common cause structure and the chain structure are Markov equivalent classes and cannot be distinguished. Formalize as:

\begin{equation} \label{9}
\rho_{xy|z}=0 \iff x \perp \!\!\! \perp y | z
\end{equation}

Finally, prior to initiating our method, it is possible to determine whether the distribution is Gaussian by examining whether the value of the third-order cumulant is zero. Subsequently, the appropriate method can be applied. As a result, our approach has unified all linear cases under a single framework.

\bibliographystyle{unsrt}
\bibliography{myreference}

\end{document}